\documentclass[runningheads]{llncs}

\usepackage{microtype}
\usepackage{graphicx}
\usepackage{subfigure}
\usepackage{float}
\usepackage{booktabs} % for professional tables
\usepackage{physics}
\usepackage{blochsphere} 
\usepackage{amssymb}
\usepackage{hyperref}
\PassOptionsToPackage{hyphens}{url}
\usepackage{microtype}
\usepackage{tikz}
\usetikzlibrary{quantikz}

\usetikzlibrary{automata,positioning}

\newcommand{\be}{\begin{eqnarray}}
\newcommand{\ee}{\end{eqnarray}}

\begin{document}

%\twocolumn[
%\icmltitle{
\title{A Review of Pulse-Coupled Neural Network Applications in Computer Vision \\and Image Processing}
\titlerunning{PCNN Applications in Computer Vision and Image Processing}

\author{Nurul Rafi
\inst{1}
\and
Pablo Rivas
\inst{2}
\orcidID{0000-0002-8690-0987} 
}

\institute{
School of Engineering and Computer Science \\
Department of Computer Science \\
Baylor University, Texas, USA\\
$^1$\email{Nurul\_Rafi1@Baylor.edu}
$^2$\email{Pablo\_Rivas@Baylor.edu}
}
\maketitle
\begin{abstract}
Research in neural models inspired by mammal's visual cortex has led to many spiking neural networks such as pulse-coupled neural networks (PCNNs). These models are oscillating, spatio-temporal models stimulated with images to produce several time-based responses. 
This paper reviews PCNN's state of the art, covering its mathematical formulation, variants, and other simplifications found in the literature. We present several applications in which PCNN architectures have successfully addressed some fundamental image processing and computer vision challenges, including image segmentation, edge detection, medical imaging, image fusion, image compression, object recognition, and remote sensing. Results achieved in these applications suggest that the PCNN architecture generates useful perceptual information relevant to a wide variety of computer vision tasks.
\end{abstract}

\section{Introduction}
\begin{figure*}[t]
\begin{center}
\includegraphics[width=\textwidth]{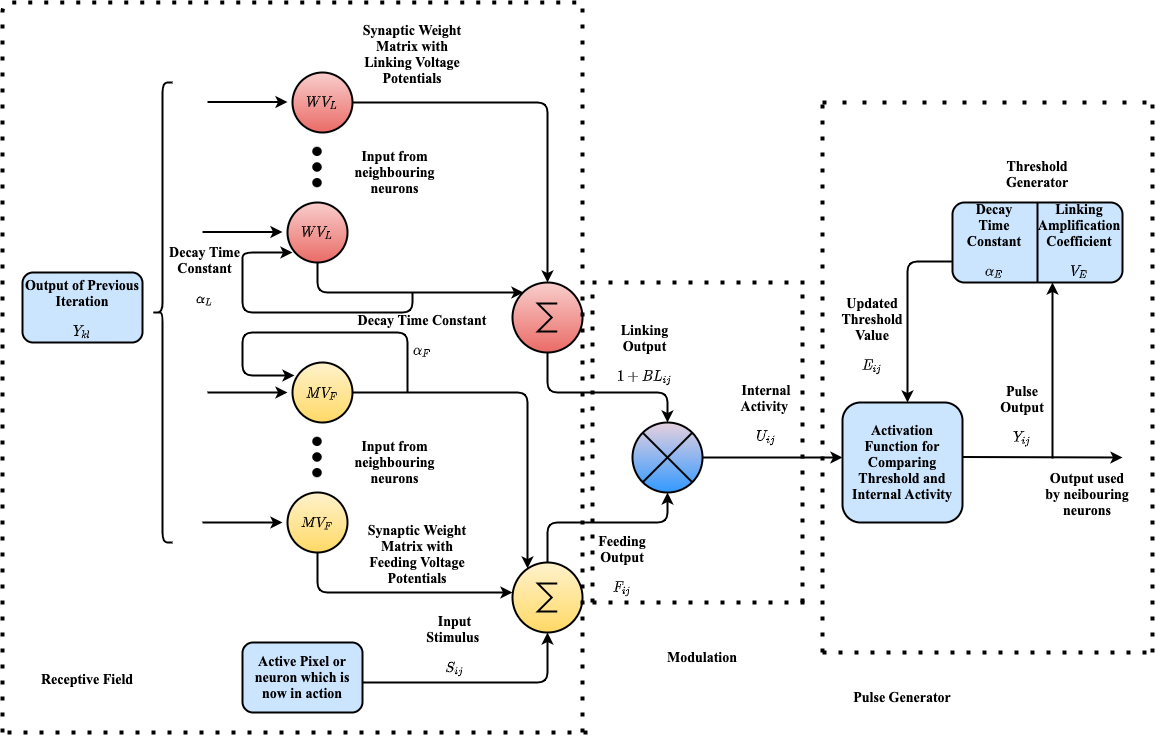}
\end{center}
   \caption{Basic Structure of a Pulse-Coupled Neural Network.}
\label{fig:pcnn}
\end{figure*}

Pulse-Coupled Neural Networks (PCNNs) belong to the category of neural networks \cite{zhan2009new,ghosh2009spiking}. This kind of neural network implements natural, biological neural models. PCNNs, in particular, implement a model of the visual cortex initially conceived by Eckhorn \emph{et al.} in the late 1980s \cite{eckhorn1989feature}. A PCNN in particular implements a mechanism by which the visual cortex of some mammals functions proposed by Eckhorn with a slight improvement that accounts for a better synchronization among neural units \cite{lindblad2005image}. The entire model comprises different mathematical operations that involve differential equations to model mechanisms for sharing information among neurons and producing different attention mechanisms that change over time \cite{kinser1996simplified}. The basic model of a neuron element of a PCNN has three main modules: a dendrite tree, a linking feed, and a pulse generator \cite{lindblad2005image}. The dendrite tree includes two particular regions of the neuron element, linking and feeding. Neighboring information is weighted in through the linking mechanism, and the feeding mechanism receives the input signal information. The pulse generator module compares the internal activity, linking, and feeding activity with a dynamic threshold that evaluates the neuron's energy potential and decides if it should fire or not. Fig. \ref{fig:pcnn} illustrates the basic model of a PCNN. 

While a PCNN is derived from Eckhorn's model \cite{eckhorn1989feature}, there are other alternative models, such as Rybak's \cite{rybak1992model} or Parodi's \cite{parodi1996temporal} which also model other similar visual cortex systems. Eckhorn's model itself was inspired by the universally known work of Hodgkin–Huxley and FitzHugh-Nagumo. Today, a PCNN has regained much attention in the computer vision world as an important image processing agent. A PCNN is usually associated with tasks such as edge detection, segmentation, feature extraction, and image filtering \cite{johnson1999pcnn,kuntimad1999perfect,keller1999pulse,ranganath1995pulse,rughooputh2000spectral,chacon2007image,wang2010review}. Because of this kind of rekindled interest in PCNNs, we carried out a literature review of the model and its applications. This paper presents our literature review analysis results and discusses some of the areas in which PCNNs have been successfully implemented.

This paper is organized as follows: Section 2 introduces the mathematical background and design of a PCNN, all its variants, and discusses recommended parameter settings. Section 3 is devoted to applications of PCNNs, and it is organized with seven subsections, each corresponding to an area of computer vision. Finally, we conclude in Section 4 with remarks about our findings and comment on this model's future research directions.

\section{Pulse Coupled Neural Network Design}
% here's where you give the details of the PCNN. e.g., make a nice diagram (visual), then from there, explain every piece mathematically 

PCNN neuron models are essentially self-trained and generate binary pulse images \cite{subashini2014pulse}, with each neuron acting as a pixel for image processing. When a neuron fires in a pulse coupled neural network, the corresponding similar areas become active automatically \cite{deng2019pcnn}. Pulse was first used to explain the idea of learning biological neuron system. \cite{ranganath1995pulse}.  Echron's pulse-coupled neuron \cite{eckhorn1989feature}, which consists of a non-linear system with a variable threshold and refractory parts, is the pioneer in this area . Later, this idea was used to develop and further research on PCNN, with the main three components of PCNN being feeding, linking, and pulse generator. In earlier literature, feeding inputs were considered receptive field and linking component as modulation \cite{subashini2014pulse}; however, the pulse generator was considered a spike generator with a dynamic threshold. 

\subsection{Process and Mathematical Definitions}

Feeding is the PCNN's input portion, where each pixel of the input image is connected to a single neuron to perform the image processing operation.
PCNN can be broken down into three components, each of which can be represented as a mathematical equation. 

The following equation was proposed by the authors of \cite{subashini2014pulse} to explain the feeding mechanism: 
\begin{align} 
F_{i j}[n]= & e^{-\alpha_{F}} F_{i j}[n-1]+S_{i j} +V_{F} \sum_{k l} M_{i j, k l} Y_{k l}[n-1] 
\end{align}
A simplified version of this, such as the following 
\begin{align} 
F_{i j}[n]=S_{i j}
\end{align}
where $i,j$ are the indices of an image of coordinates $I[i,j]$, and $n$ is the $n$th neural unit, since each pixel of input is treated as a separate neuron. Here, $F_{i,j}$ is the initial feeding input \cite{subashini2014pulse}, and $S_{i,j}$ is the normalized pixel gray value input to the neuron. The neighbouring pixels corresponding to the current active pixel are represented by $kl$, and the weight matrix is represented by $M$ \cite{deng2019pcnn}. $Y_{kl}$ is the output from previous iteration and the exponential decay factor is defined by $e^{-\alpha_{F}}$ \cite{wang2020improved}.

Then there's the linking mechanism:
\begin{align}
    L_{i j}[n]=e^{-\alpha_{L}} L_{i j}[n-1]+V_{L} \sum_{k l} W_{i j, k l} Y_{k l}[n-1]
\end{align}
where the simplified version is:
\begin{align}
L_{i j}[n]=V_{L} \sum W_{i j k l} Y_{k l}[n-1]
\end{align}

We've already established that the receptive field is made up of two subsystems: a linking subsystem and a feeding subsystem. So, in the receptive area $L i,j$ is the contribution of the linking subsystem \cite{deng2019pcnn}. The connecting weight matrices inside other neurons are $W$ and $M$ from the feeding compartment \cite{subashini2014pulse}. 
Their job is to link nearby neurons to the center neuron, which is currently active \cite{ganasala2016feature}. In addition, $W_{ij,kl}$ can be represented as follows:
\begin{align}
W_{i j, k l}=\frac{1}{\sqrt{(i-k)^{2}+(j-l)^{2}}}
\end{align}
which is the Euclidean distance between eight neighbor neurons. \cite{he2019parameter}. The action is then followed by the linking or modulation portion. The modulation equation is as follows:
\begin{align}
U_{i j}[n]=F_{i j}[n]\left[1+\beta L_{i j}[n]]\right.
\end{align}

The total internal activity of that specific neuron that has come from the feeding and linking subsystems is $U_{i,j}$. It makes up the modulation component \cite{subashini2014pulse}. The modulation subsystem's linking coefficient, $\beta$ \cite{deng2019pcnn}, specifies how many pixels are coupled with the surrounding pixels or neurons \cite{xu2014multiplicative}.

Finally, there's the pulse generator, which includes a threshold generator and an activation function. The following equation can be used to define the first threshold value:
\begin{align}
E_{i j}[n]=e^{-\alpha_{E}} E_{i j}[n-1]+V_{E} Y_{i j}[n-1]
\end{align}

If the threshold value is exceeded, the neuron fires. After a neuron fires, the threshold value begins to decay before the neuron fires again for the next iteration.
This decay happens to regulate the neuron's ability to fire again \cite{wei2011automatic}. Until firing again, the value of internal activity will decay exponentially before it reaches this threshold value again, and this interim time is known as the refractory period \cite{johnson1999pcnn}. The output is fed back to the threshold generator, which dynamically changes the threshold and still checks for the output with the latest internal activity, $U$. When the threshold is greater than $U$, output will gradually become zero \cite{subashini2014pulse}.

Finally, based on the threshold value, the final output will be as follows:
\begin{align}
Y=\left\{\begin{array}{ll}
1, & \text { if } U_{i j}[n]>E_{i j}[n] \\
0, & \text { otherwise }
\end{array}\right\}
\end{align}

\begin{figure*}[t]
\begin{center}
\includegraphics[width=\textwidth]{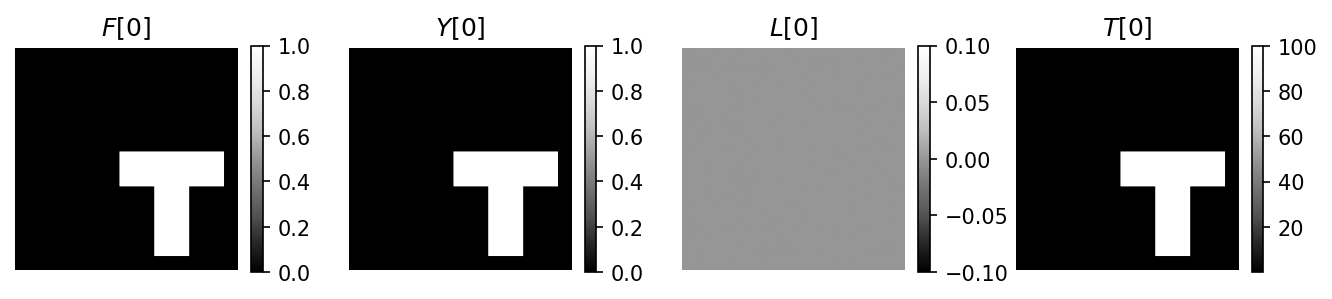}
\includegraphics[width=\textwidth]{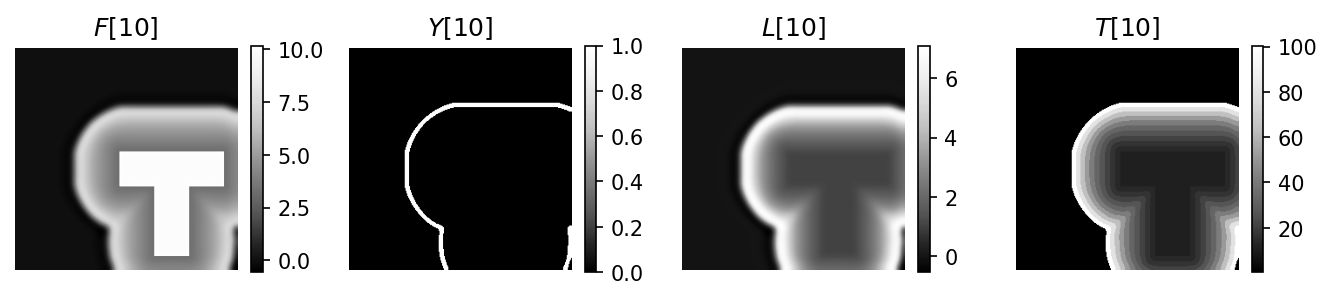}
\includegraphics[width=\textwidth]{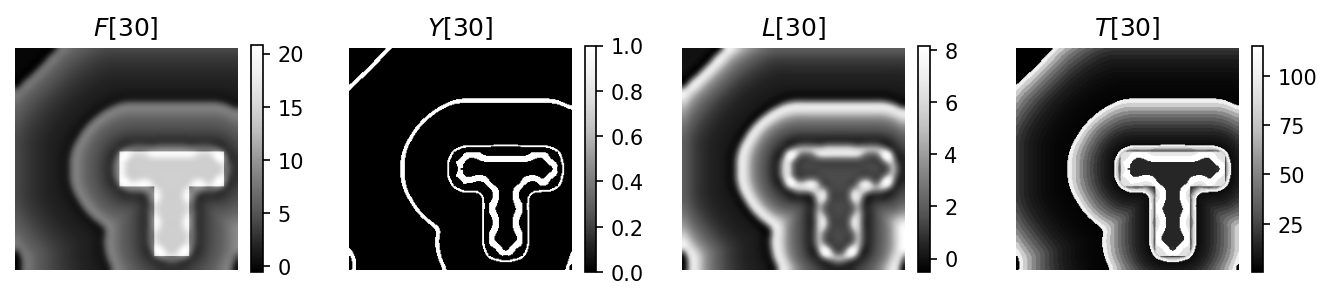}
\end{center}
   \caption{Examples of pulses of PCNN. The top row is the initial iteration, the middle row is the 10-th pulsation, and the bottom row is the result after 30 pulses.}
\label{fig:pcnneg}
\end{figure*}

PCNN produces a sequence of pulse outputs after $n$ iterations, which can be analyzed to make decisions about the input image. Fig. \ref{fig:pcnneg} illustrates an example of such pulses at various times. The actual performance of the network is $Y[n]$, as seen in the figure, but the rest of the elements are included for comparison. 

Here $V_F$, $V_L$ are the inherent voltage potentials \cite{subashini2014pulse} and $V_E$ is the linking amplification coefficient between the output and threshold generator \cite{deng2019pcnn}. Also, $\alpha_{F}$, $\alpha_{L}$, $\alpha_{E}$ are the time constants for the iteration decay for the related subsystem and determine the internal status of the network \cite{zhou2016simplified}.

The inherent voltage potentials are $V_F$ and $V_L$ \cite{subashini2014pulse}, and the linking amplification coefficient between the output and threshold generator is $V_E$ \cite{deng2019pcnn}. Also, $\alpha_{F}$, $\alpha_{L}$, $\alpha_{E}$ are the time constants for the iteration decay for the associated subsystem, and they decide the network's internal status \cite{zhou2016simplified}.

\subsection{Parameter Settings}
Since the PCNN model has so many parameters, some of them must be initialized before the PCNN can perform. These parameters have a direct impact on PCNN's results. Finding automated parameter settings is still a difficult job. the authors of this paper\cite{wei2011automatic} introduced automatic adjustment of threshold decay constant. The linking coefficient $beta$ was automated in this article \cite{kuntimad1999perfect}. The automatic parameter settings for PCNN for image segmentation were introduced by connecting the neurons and the input image \cite{chen2011new}.  Generally, $\alpha_F < \alpha_L$ \cite{wang2010multi} and and $\alpha_E$ is always less than 1 \cite{zhu2017memristive}.

Initially \cite{berg2008automatic} and \cite{ma2006study} introduced PCNN's automated parameter settings. They suggested a new condensed version of SPCNN based on SCM because PCNN requires intensive training \cite{zhan2009new}. SCM outperformed regular PCNN due to its lower time complexity and the fact that SPCNN needs no previous training. They adjusted 5 parameters in total: $\alpha_F, \alpha_L, V_E, V_L, \beta$ were used to establish a relationship between dynamic neurons and the input image \cite{chen2011new}. They came up with the following equation for $alpha_F$:
\begin{align}
\alpha_{F}=\log \left(\frac{1}{\sigma_I}\right)
\end{align}
where $sigma_I$ is the standard deviation of the input image $I$, whose amplitude has been normalized. 

The key parameters for efficient image segmentation are the linking coefficient, $beta$, and the exponential decay factor, $alpha E$ \cite{subashini2014pulse}. These two parameters are both in charge of detecting image edges \cite{deng2014pcnn}. Image segmentation efficiency can be influenced by decay factors. When segmenting images, different values of $alpha_E$ produce better results, but a higher value produces a bad result. In addition, a variable and small $beta$ performs better than a fixed one that maintains synchronous pulse \cite{zhou2016simplified}. $W$ and $V_E$ are also crucial parameters for improving PCNN efficiency. Authors of this paper attempted to automate the decay time constant, $alpha E$ by using the following equation for segmentation,  which outperformed Otsu and $K$-means methods \cite{wei2011automatic}.
\begin{align}
\alpha_{E}=C / \mu
\end{align}
where $C$ is a constant and $\mu$ is the average of the input image's grey level.
$V_E$ is a broad value that affects the firing time of neurons.
Individual applications also affect $alpha_F$, $alpha_L$, and $alpha_E$.  \cite{yang2019overview}.

It's difficult to find the right set of parameters for a PCNN since it depends on the application. The parameters shown in Table \ref{tbl:aTable} are the recommended ones, as found in the literature, for most applications that require a stable version of the PCNN. 
\begin{table}[t]
 \caption{Recommended PCNN Parameter Setting}
 \label{tbl:aTable}
 \begin{center}
  \begin{tabular}{|c|c|c|c|}
    \hline 
    Param   & Recommended & Source & Model\\
    \hline \hline
    $V_L$ & 1 & \cite{zhou2016simplified} & S-PCNN\\
     & 0.01 & \cite{helmy2016image} & SOM-PCNN\\
     & 0.5 & \cite{yang2020combined} & HMM-PCNN\\
    $V_F$ & 0.5 & \cite{yang2020combined} & HMM-PCNN\\
     & 1 & \cite{zhou2016simplified} & S-PCNN\\
    $V_E$ & 0.0001 - 400 & \cite{he2019parameter} & SPCNN-Cuckoo\\
     & 10 & \cite{helmy2016image} & SOM-PCNN\\
     & 400 & \cite{wang2020improved} & SPCNN-Intensity\\
     & 20 & \cite{wang2018novel} & PCNN-Random\\
     & 20 & \cite{wang2016leaf} & PCNN-SVM\\
    $\alpha_E$ & 0.0001 - 100 & \cite{he2019parameter} & SPCNN-Cuckoo\\
     & 0.089 & \cite{helmy2016image} & SOM-PCNN\\
     & 0.075 & \cite{ganasala2016feature} & SPCNN-NSST\\
     & 0.2 & \cite{wang2016leaf} & PCNN-SVM\\
    $\beta$ & 0 - 1 & \cite{xu2014multiplicative} & PCNN-Factoring\\
     & 0.0001 - 100 & \cite{he2019parameter} & SPCNN-Cuckoo\\
     & 0.2 & \cite{helmy2016image} & SOM-PCNN\\
     & 2-0.1 & \cite{wang2020improved} & SPCNN-Intensity\\
     & 0.1 & \cite{wang2016leaf} & PCNN-SVM\\
    \hline 
  \end{tabular}
 \end{center}
\end{table}

These are the empirical values. However, setting parameters automatically is a difficult job as well. Authors attempted to use Shannon and cross-entropy to arrive at the best threshold value \cite{yi2004automated,ma2002automated}. Szekely is the first to discover the adaptive network parameters for PCNN in this area \cite{szekely1999parameter}. 

\section{Applications of PCNN}
PCNN has a wide range of applications, especially in computer vision. \cite{lindblad2005image}. PCNN is widely used in fields such as segmentation, fusion, feature and edge detection, noise reduction and pattern recognition, and medical research \cite{subashini2014pulse}. The applications of PCNN will be discussed in depth here.

\subsection{Image Segmentation}
Where multiple regions can be identified in an input image, image segmentation identifies the most similar region based on the same characteristics like intensity or texture. Image segmentation is a technique for identifying objects by grouping pixels in a specific region. Object detection is started with this concept. Based on unit linking, they described PCNN as a self-organized and efficient method for segmenting various digital images \cite{gu2002new}. The conventional and true approach is based on the grayscale segmenting threshold \cite{yi2004automated}.

For image segmentation or other image processing tasks, unit linking PCNN works in parallel while mathematical morphology works in sequence, which is faster than conventional operations \cite{gu2005general}. It's a difficult job to segment various types of images using different types of PCNN parameters settings. This unit linking PCNN for automatic segmentation for different types of images without setting different parameters for the model was introduced by XIAO-DONG GU~\cite{gu2002new}.

Processing edge pixels is not simple enough for multi-value image segmentation.
PCNN generates small regions called seeds, and each pixel corresponds to a seed, resulting in a final matrix of multiple-segmented regions \cite{lu2008new}. PCNN uses cross-entropy for image segmentation, calculating the two input and segmented output images per iteration. Finally, the best output is observed when the cross-entropy is minimized \cite{yi2004automated}.

PCNN overcomes much of the drawbacks of other image segmentation approaches, which take longer time and are less accurate \cite{subashini2014pulse}. To change the connection strength coefficient in the model, they used a smaller number of parameters in this paper \cite{wang2010multi}. Researchers in this paper used bidirectional search to solve a color imaging problem using all of the image's details, which solved low-speed computing \cite{wang2012palmprint}. In this article, path multi-object segmentation is used~\cite{song2010one}.

Researchers developed ICS-PCNN, which uses an improved cuckoo search algorithm for human infrared segmentation to increase convergence speed and search performance. The ICS-PCNN model outperformed other PCNN variants and different segmentation models in a study of 100 infrared images \cite{he2019parameter}. The internal operation has been simplified for better segmentation in the SM-ICPCNN version \cite{chen2011new}. They reported that their approach is far superior to other models for medical, color, and grayscale images. This model had a higher overlap rate, robustness, sensitivity, precision, and Area under the Curve (AOC) for the majority of the images \cite{yang2018saliency}.

A self-organizing map (SOM) is used in conjunction with a modified PCNN to reduce classification error, particularly over-segmentation. The input using spatial frequency undergoes the most significant change in this case. Though segmentation is a difficult process, particularly for high-resolution images, for segmentation, the combination of SOM and modified PCNN outperformed other models such as Fuzzy c-means, convex relaxed kernels, and others \cite{helmy2016image}. 

With automated parameter settings, SPCNN outperformed standard PCNN and the normalized cur method \cite{shi2000normalized} in segmentation with high contrast images rather than lower contrast images \cite{chen2011new}.

\begin{align}
S_{i j}=\sum_{i=1}^{M} \sum_{j=1}^{N}\left(I_{i j}-I_{i-1, j}\right)^{2}+\left(I_{i j}-I_{i, j-1}\right)^{2}
\end{align}
\begin{align}
F_{i j}[n]=\text { normalized SF } S_{i j}
\end{align}

Using Maximum Shannon entropy and Maximum variance ratio, different components of grayscale images were used to produce color features from the input images. The proposed method preserved the texture, edges, and brightness of the input color image, despite the lengthy processing time required by the segmentation graph's large number of iterations \cite{li2018color}.

\subsection{Edge Detection}
Salient object detection \cite{shi2015hierarchical} plays an important role in image segmentation \cite{gu2005general,luo2017object,kinser1996simplified}, feature extraction \cite{jiang2013salient} more than semantic segmentation \cite{wang2020improved}. In different types of dataset the proposed SPCNN performed better than existing other seven methods: SRM, NLDF, C2S, DSS, AMU, DGRL, PiCaNet-R. They used pixel intensity as the parameters instead of usual network parameters \cite{wang2020improved}.

The following equation was used for linking input in the Simplified Region Increasing PCNN (SRG-PCNN), which can handle edge pixels very well despite the model's higher time complexity \cite{lu2008new}.
\begin{align}
L_{i}[t]= & \operatorname{step}\left(\sum_{z \in N(i)} Y_{z}[t]\right)  =\left\{\begin{array}{ll}
1 & \text { if } \sum_{z \in N(i)} Y_{z}[t]>0 \\
0, & \text { otherwise }
\end{array}\right.
\end{align}

\subsection{Medical Imaging}
Automatic segmentation and classification in dentistry is difficult due to the complex arrangement of teeth. Gaussian filtering regularized level set (GFRLS) and improved PCNN outperformed Fuzzy c-means clustering \cite{kannan2012effective}, Density based spatial clustering(DBSCAN) \cite{ester1996density}, heiarchical cluster analysis (HCA) \cite{ghebremedhin2015validation} and Gaussian Filter models(GSM)  \cite{mcnicholas2010serial}. They used MicroCT images as the dataset, and improved PCNN could classify using all of the details and the resulting hierachical images. This is their improved version of PCNN:
\begin{align}
\xi(n) & = U[n]-E[n] \\
G_{i j} & = \sum_{r} \sum_{t}\left|\xi_{i j}[n]-\xi_{i+r, j+t}[n]\right| \\
Y_{i j} & = \left\lceil\frac{\xi_{i j}[n]}{\max \xi[n]} \times k\right\rceil,
\end{align}
where G denotes the variance between internal activity and the threshold value.
Also, output is influenced by normalization multiplied by the $k$ parameter, which represents MicroCT image hierarchies \cite{wang2016automatic}.

The role of medical image fusion is becoming increasingly relevant as the number of imaging clinical applications grows. Researchers developed m-PCNN, a modern multi-modal channel that overcomes the limitations of conventional PCNN and outperforms other current fusion methods, where m is the number of input channels. They believed that their approach worked better with more information in the input images than other methods, and that it could benefit doctors more \cite{wang2008medical}.

For image fusion, modified PCNN and nonsubsampled shearlet transform (NSST), which has high directional sensitivity and low computational complexity, are used with CT-MRI and SPECT-MRI datasets \cite{easley2008sparse}. The input is shifted into low and high frequency bands of different scales using NSST, and the bands are combined using PCNN fusion to produce fused LF. Image fusion is used to obtain more accurate information from the source image using multimodality of medical images \cite{ganasala2016feature}.

To solve the various optimization problems for multi-modal medical images, the researchers combined a modified PCNN with a quantum behaved particle swarm optimization (QPSO) algorithm. They evaluated three parameters to find a good fitness function: image entropy (EN), average gradient (AG), and spatial frequency (SF) \cite{xu2016multimodal}. They devised the multi-criteria fitness function as follows:
\begin{align}
f=\max (S F+E N+A G).
\end{align}

Breast cancer is a frightening disease for women, and the number of patients using mammograms for screening is rising every day all over the world \cite{lee2010breast}. It causes a problem to check manually because the poor contrast between the lesion and normal tissues makes it more difficult to examine mammogram images. Researchers implemented PCNN with a level set system for breast cancer screening, using the MIAS database to measure breast masses. Boundary leakage and unwanted context segmentation are avoided using the level set approach in this case \cite{xie2016pcnn}.

M-PCNN is a memristive pulse coupled neural network for medical image fusion that has a stronger fusion effect while preserving focus consistency. Input image noise can be reduced by adjusting the brightness of the pixel M-PCNN. Edge detection and extraction are also possible with M-PCNN using gray mutation of the edge \cite{zhu2017memristive}. Gray correlation is used with PCNN instead of weight matrix or Euclidean distance between pixels for image segmentation, though the running time is longer than other approaches. Another issue was that their proposed approach couldn't take advantage of the entire digital input space and comprehensive information \cite{ma2014automatic}.

\subsection{Image Fusion}
Image fusion is the process of combining multiple image sources into a single unified format that contains more details \cite{behrenbruch2003fusion}. Image fusion also saves money on storage since it combines many images into a single file \cite{xu2016multimodal}. In this paper \cite{johnson1999pcnn} they presented PCNN as a potential in the field of image fusion.

Multi-focus image fusion combines multiple images with different focus settings into a single image. DWT \cite{li1995multisensor} with wavelet and gradient pyramid \cite{burt1992gradient} methods were previously used for multi-focus image fusion, but they are complex and time consuming. Instead of using simple PCNN for image fusion, researchers suggested using dual-channel PCNN for better performance and quality. Instead of using normal PCNN, dual-channel PCNN handles the weighting factor based on the weighting coefficients and two input channels to do well with multi-focus fusion~\cite{wang2010multi}.

To obtain high-frequency information from the input images, redundant lifting non-separable wavelet multi-directional analysis (NSWMDA) for decomposition in different subbands is combined with modified PCNN.
To obtain the ultimate fused image of directional sub-bands, Gaussian sum-modified-Laplacian (GSML) is combined with PCNN \cite{zhao2014image}. 

For automatic parameter settings of simplified PCNN, particle swarm optimization (PSO) is used, and image fusion is done by creating sub-blocks from the input image.
In image fusion with multi-focus input image, this proposed method performed better than PCA, SIDWT, and FSDP.
 \cite{jin2018multi}.

Shearlet can perform Multi-scale Geometric Analysis (MGA) as well as rich mathematical analysis for multidimensional data, similar to how wavelet does for one-dimensional data.
For image fusion, the proposed PCNN used multi-scale and multi-directional image decomposition and outperformed other approaches in terms of extracting more accurate information from optical and SAR images~\cite{cheng2013novel}.

The random walk model distinguishes between objects with similar luminance, reduces noise, and detects the targeted area effectively.
It performed admirably in image fusion for multi-focus input images, alongside PCNN \cite{wang2018novel}.

A Surfacelet transform, when combined with PCNN, is a strong multi-resolution tool that outperforms conventional image fusion methods.
Compound PCNN was proposed with local sum-modified Laplacian to solve the disadvantages of standard PCNN, where dual channel PCNN was combined with PCNN to find the fusion coefficients.
In image fusion, this combined PCNN outperformed PCA, DWT, and LAP methods using the shared knowledge (MI) and QAB/F metrics~\cite{zhang2014multi}.

Researchers used PCNN for image fusion from multi-focus images after adjusting the PCNN parameters to account for sharpness.
They concentrated on automating the setting of the $beta$ parameter, which is the pixel linking coefficient~\cite{miao2005novel}. The proposed model for efficient image fusion is non-subsampled shearlet transform (NSST)–spatial frequency (SF)–pulse coupled neural network (PCNN), where NSST has lower time complexity than the MGA tool.
In shift-invariance, multi-scale, and dimensional image results, NSST performs better~\cite{kong2014novel}.

\subsection{Image Compression}
In the case of high-dimensional data, multi-scale geometric analysis outperforms the wavelet transform approach, and the contourlet transform is the preferred approach for orientation-based image coding \cite{do2006contourlet}. The HMM-contour model worked better at removing redundant information and extracting crucial information from the image PCNN. The coefficients from the input image are obtained using the contourlet transform, which is subsequently converted to a tree structure by HMM. 

The HMM parameters are estimated using the EM algorithm. Furthermore, the SPIHT algorithm is used for coding and transmitting PCNN-classified subband coefficients \cite{yang2020combined}. In this paper they proposed set partitioning coding system(SPACS) which performs better than SPIHT algorithm \cite{li2015generalization}.

\subsection{Object Recognition}
PCNN with region-based object recognition simplified
Chen \cite{chen2014region} introduces SPCNN-RBOR, which uses primarily color image segmentation for object detection and Scale-invariant feature transform (SIFT) \cite{lowe2004distinctive}, which is common in object recognition for its accuracy and speed.
In object detection based on textures, their proposed method outperformed and overcome the drawbacks of feature-based methods. In object detection based on textures, their proposed method outperformed and overcome the drawbacks of feature-based methods \cite{chen2014region}.

Support Vector Machine (SVM) was used as a classifier for different types of leaves in this study \cite{wang2016leaf}, while PCNN was used for leaf recognition.
To obtain the classification, texture and shape information are extracted from the input image. With improved PCNN, crack detection in metal bodies is easy, and crack spots can be detected with effective threshold selection.
The image was tested for small cracks using a magnetic, optic image (MOI) calculated by a CCD sensor \cite{cheng2018research}.

\subsection{Remote Sensing}
The most important remote sensing knowledge for Baltic Sea ice is synthetic aperture radar (SAR) images.
To identify and segment SAR images, a modified PCNN is used, with the error level determined using the Gaussian distribution function.
In this case, maximum resolution (100 m) Radarsat-1 ScanSAR Broad mode images were used, and PCNN performed well in terms of classifying images in terms of execution time \cite{karvonen2004baltic}.

For land observation and environmental change, analyzing change detection from remote sensing images is important \cite{yetgin2011unsupervised}. For high-resolution HSR remote sensing image observation, MPCNNCD-Modified PCNN change detection is proposed, as HSR contains more spatial information than other forms of remote sensing images \cite{zhao2014image}. MPCNN employs a normalized moment of inertia (NMI) function to effectively detect hot spot areas.
Finally, the final map of hot spot areas is generated using the expectation-maximization (EM) algorithm \cite{dempster1977maximum} \cite{zhong2014change}.

\subsection{Noise Removal}
Noise reduction is an important method for getting better results from input images, and PCNN can helps with this because it makes efficient use of the network.
When there is a mismatch sequence in the neighboring pixel of an image, noise occurs. To remove salt and pepper noise, PCNN was combined with a median filter but it was not enough capable of removing Gaussian noise \cite{subashini2014pulse}.
Gaussian noise was eliminated using PCNN and the Median and Weiner filters \cite{yi2003gaussian}. In this paper \cite{xiong2010analog} they proposed a method which is better than Median, Lee and Weiner filter.

Researchers suggested a simpler PCNN with Median filter, which performed better not only in noise removal but also in maintaining the image's originality, due to the high complexity of the Median filter for removing impulse noise of an image \cite{yi2003new}.

\section{Conclusions}

This paper reviews the state of the art concerning pulse-coupled neural networks. We covered its mathematical formulation, variants, and simplifications found in the literature. Then we presented seven applications in which PCNN architectures are successful in addressing image processing and computer vision tasks. These include image segmentation, edge detection, medical imaging, image fusion, image compression, object recognition, and remote sensing. These applications' results suggest that the PCNN architecture may be considered a funcional pre-processing element to increase vision systems' performance. The findings demonstrate PCNNs ability to generate useful perceptual information relevant to a wide variety of tasks. Note that most of these tasks are highly complex, and the results are unique in each case.

There are some clear opportunities for research in PCNNs, which we can summarize as follows:
\begin{itemize}
    \item {\it Computational cost} seems to be a problem if we compare PCNNs to other traditional image processing techniques. More research is needed to optimize the model and its parameters for maximum performance.
    \item It is not clear if there is an {\it ideal machine learning method} that can be paired with a pre-processing PCNN, particularly if the method can be naturally paired or introduced as part of the PCNN to extend its capabilities. For example, using fuzzy set theory or support vector machines.
    \item While much research focuses on using a PCNN by setting its parameters to produce stability, there is not enough work on {\it exploiting the chaotic behavior} that is available for exploration. We do not know if there is a fitting application of such chaotic neural behavior in some, a group, or in all of the neurons in a PCNN.
    \item {\it Automatic parameter setting} is a well-known problem that has been partially solved for some simplified versions of the PCNN but no for the standard PCNN.
    \item We need to incorporate more research and recent {\it advances in neuroscience and neurobiology} into the PCNN to improve its formulation. 
\end{itemize}

%\section*{Acknowledgements}
%The authors thank the **omitted for blind review** for their support under grant **omitted for %blind review**. This research was also funded, in part, by **omitted for blind review**.

\bibliographystyle{splncs04}
\bibliography{example_paper}

\end{document}